\documentclass{article}
\usepackage{tikz}
\usepackage{amsmath,amssymb,amsfonts}

\usepackage{graphicx}

\usepackage{tikz}
\usetikzlibrary{arrows.meta,positioning,calc,fit,backgrounds,decorations.pathreplacing}
\definecolor{accent}{HTML}{2563EB}   % trainable
\definecolor{frozen}{HTML}{9CA3AF}   % frozen

\tikzset{
  >={Latex[length=2mm]},
  font=\sffamily\small,
  box/.style={draw=black, rounded corners=2pt, very thick, fill=white, align=center, minimum width=34mm, minimum height=14mm},
  thinbox/.style={draw=black, rounded corners=2pt, thick, fill=white, align=center, minimum width=28mm, minimum height=10mm},
  trainable/.style={draw=accent, very thick},
  frozenstyle/.style={draw=frozen, dashed, very thick},
  edge/.style={-Latex, very thick},
  tiny/.style={font=\sffamily\footnotesize, text=gray!70},
  labeltick/.style={font=\sffamily\scriptsize, text=gray!70},
}

\usepackage[english]{babel}

\usepackage[letterpaper,top=2cm,bottom=2cm,left=3cm,right=3cm,marginparwidth=1.75cm]{geometry}
\usepackage{pgfplots}
\pgfplotsset{compat=1.18}

\usepackage{amsmath}
\usepackage{graphicx}
\usepackage[colorlinks=true, allcolors=blue]{hyperref}

\title{Foley Control: Aligning a Frozen Latent Text-to-Audio Model to Video}

\author{
  Ciara Rowles, Varun Jampani, Simon Donn\'{e}, Shimon Vainer,\\
  Julian Parker, Zach Evans \\[4pt]
  \textit{Stability AI}
}

\begin{document}
\maketitle
% --- TikZ + styles (preamble) ---

\begin{abstract}
Foley Control is a lightweight approach to video-guided Foley that keeps pretrained single-modality models frozen and learns only a small cross-attention bridge between them. We connect V-JEPA2 video embeddings to a frozen Stable Audio Open DiT text-to-audio (T2A) model by inserting compact video cross-attention after the model’s existing text cross-attention, so prompts set global semantics while video refines timing and local dynamics. The frozen backbones retain strong marginals (video; audio given text) and the bridge learns the audio–video dependency needed for synchronization --- without retraining the audio prior. To cut memory and stabilize training, we pool video tokens before conditioning.
On curated video–audio benchmarks, Foley Control delivers competitive temporal and semantic alignment with far fewer trainable parameters than recent multi-modal systems, while preserving prompt-driven controllability and production-friendly modularity (swap/upgrade encoders or the T2A backbone without end-to-end retraining). Although we focus on Video-to-Foley, the same bridge design can potentially extend to other audio modalities (e.g., speech).\end{abstract}

\section{Introduction}

Sound design is central to immersion in film, games, and VR: subtle contact sounds, material cues, and timing-sensitive transients anchor visual events in a coherent perceptual scene. Although recent video-to-audio (V2A) systems have advanced fidelity and semantic coverage, they frequently entail heavy training pipelines or control stacks that limit practicality in production settings~\cite{cheng2024mmaudio,shan2025hunyuanfoley,wang2025klingfoley}.

Broadly, prior work splits into two paths. Adapter-based methods (e.g., FoleyCrafter~\cite{zhang2024foleycrafter}) plug semantics and timing controllers into strong T2A generators, improving alignment without retraining large backbones. By contrast, end-to-end foundation V2A models demand far more data with tightly aligned video–audio pairs to learn the audio prior, cross-modal mapping, and temporal synchrony simultaneously — driving curation of massive paired corpora with heavy filtering (onset heuristics, CLAP/ImageBind screening, alignment checks) and representation-alignment losses~\cite{cheng2024mmaudio,shan2025hunyuanfoley}. Scale is further hampered by real-world noise: dubbing, off-screen sources, background music, and imprecise timestamps degrade supervision and underrepresent long-tail events. We instead freeze a strong T2A backbone and learn a thin video–audio bridge, attaining competitive alignment with far less data: our corpus uses \textasciitilde{}700k Kinetics\textendash{}700 clips, whereas HunyuanVideo\textendash{} Foley trains on \textasciitilde{}100k hours (\(\approx 3.0{\times}10^{7}\) twelve\textendash{}second segments), i.e., \(\sim 43{\times}\) more paired data.

At the other extreme, multimodal diffusion transformers (e.g., MMAudio~\cite{cheng2024mmaudio}, HunyuanVideo-Foley~\cite{shan2025hunyuanfoley} and others ~\cite{chen2025video} \cite{tian2025audiox} \cite{karchkhadze2025stereofoley}) jointly train audio, video, and sometimes text streams end-to-end.
Such approaches achieve impressive synchronization and coverage, but at the cost of massive curated datasets, high compute budgets, and reduced modularity.

This paper proposes \textbf{Foley Control}, a lightweight framework that targets the same alignment benefits while preserving the practicality of frozen generative backbones. Our key idea is to connect V-JEPA2~\cite{assran2025vjepa2} video embeddings to a frozen Stable Audio Open DiT~\cite{evans2024stableaudioopen} by inserting \emph{collaboration layers}---compact, video-conditioned cross-attention modules placed \emph{inside} existing transformer blocks. This placement is deliberate: video cross-attention is applied \emph{after} the model’s original text cross-attention, so text prompts first establish high-level semantics and structure, and video then refines temporal grounding and localized dynamics. By freezing the remaining parameters of the DiT blocks, we retain the strong generative prior learned from large-scale audio--text corpora and focus the trainable capacity on cross-modal synchronization rather than relearning audio generation.

From an architectural perspective, we employ a streamlined integration strategy in which V-JEPA2 embeddings are pooled into a compact grid representation and injected through lightweight cross-attention modules placed inside the frozen DiT blocks. This design ensures that video information refines the audio latent trajectory without altering the established text-conditioning pathway. Rotary position embeddings (RoPE)~\cite{su2021roformer} further enhance temporal grounding by providing ordering signals across modalities, eliminating the need for heavier synchronization mechanisms. The resulting architecture remains compact yet expressive, scaling effectively to longer contexts and diverse scenes while preserving prompt-driven controllability.

Taken together, these elements provide a practical route to high-quality Foley generation: reuse a strong, frozen T2A backbone for audio fidelity and prompt control, and add just enough trainable capacity to align timing and dynamics to the video.

\section{Related Work}

\paragraph{FoleyCrafter.}
Early neural Foley systems learn to synthesize sounds that are semantically and temporally aligned with visual inputs, but often depend on limited audio--visual data and struggle to preserve high audio fidelity.
FoleyCrafter~\cite{zhang2024foleycrafter} addresses this by \emph{plugging} lightweight controllers into a strong text-to-audio backbone, thereby retaining audio quality while improving video--audio alignment.
Concretely, it builds on a U-Net–based V2A generator (in the spirit of AuFusion~\cite{xue2024auffusion}) and employs \emph{multiple} control streams: a \emph{semantic adapter} that injects video/text features throughout the U-Net (early, middle, and late blocks), and a \emph{timestamp/onset controller} that is applied primarily in late layers to sharpen synchronization around transient events.
Event timing cues are provided by a \emph{separate} timestamp detection model , whose outputs modulate the diffusion steps to align onsets without altering the pretrained backbone.
This division of labor—frozen backbone for fidelity, semantic control across the network, and late-layer timing refinement—yields stronger alignment under modest compute.

\paragraph{MMAudio.}
MMAudio~\cite{cheng2024mmaudio} introduces a unified, \emph{from-scratch} multimodal training paradigm that jointly leverages audio--text and audio--video pairs under a conditional flow-matching objective.
A hybrid architecture---multimodal DiT blocks followed by audio-only blocks---supports scalable data mixing and strong semantic alignment, while a synchronization module operating via high frame-rate visual features further improves temporal precision.
A related approach, HunyuanVideo-Foley~\cite{shan2025hunyuanfoley}, scales this paradigm with a massive curated text--video--audio dataset and a dual-stream multimodal diffusion transformer that fuses audio--video attention with text cross-attention.
Additionally, it introduces a representation-alignment loss (REPA\cite{yu2025representationalignmentgenerationtraining}) that steers the audio DiT’s hidden states toward self-supervised audio embeddings, enhancing fidelity and stability, and employs a DAC-style autoencoder for higher-quality waveform reconstruction.

\paragraph{Stable Audio Open.}
Stable Audio Open~\cite{evans2024stableaudioopen} is a foundation text-to-audio model based on latent diffusion, combining a fully convolutional VAE, T5-based text conditioning, and timing embeddings to enable efficient generation of variable-length 44.1kHz stereo signals up to 95 seconds.
Despite operating in a compressed latent space, it achieves state-of-the-art fidelity on both music and sound effects, offering a strong frozen backbone for adaptation in multimodal alignment tasks.
Early text-to-audio (T2A) diffusion models such as DiffSound~\cite{yang2022diffsound}, AudioGen~\cite{kreuk2023audiogen}, AudioLDM~\cite{liu2023audioldm}, and Make-An-Audio~\cite{huang2023makeanaudio} established the latent diffusion paradigm for sound synthesis.
Stable Audio Open~\cite{evans2024stableaudioopen} extends this approach with high-fidelity 44.1\,kHz generation and strong semantic conditioning, while TangoFlux~\cite{hung2025tangoflux} explores fast flow-matching variants for text-conditioned audio generation.

\paragraph{V-JEPA2.}
V-JEPA2~\cite{assran2025vjepa2} is a large-scale self-supervised video model designed to learn predictive representations of the physical world from internet-scale video. It extends the joint-embedding predictive architecture (JEPA) by scaling pretraining to over one million hours of video and up to one billion parameters, using a masked feature prediction objective in representation space. Unlike generative approaches that reconstruct pixels, V-JEPA2 focuses on predictable dynamics such as motion trajectories, yielding stronger representations for action understanding, anticipation, and temporal reasoning. The model achieves state-of-the-art results on motion understanding benchmarks (e.g., 77.3 top-1 accuracy on Something-Something v2) and human action anticipation (39.7 recall-at-5 on Epic-Kitchens-100), while also supporting downstream video question-answering when aligned with large language models. These properties make V-JEPA2 a compelling choice for video-conditioned generative tasks such as Foley synthesis, where fine-grained motion cues and temporal structure are critical

\paragraph{Other related V2A / audiovisual models.}  
Several recent works also explore video-to-audio or joint audiovisual generation along different tradeoffs~\cite{mo2024tta,zhang2024foleycrafter,luo2023difffoley,wang2024frieren,cheng2024mmaudio,liu2025thinksound,shan2025hunyuanfoley}.

For instance, \textbf{FRIEREN} proposes rectified flow matching in spectrogram latent space to regress a conditional transport vector field, enabling few-step or even one-step audio sampling with strong video-audio alignment \cite{wang2024frieren}.  
\textbf{UniVerse-1} fuses pretrained video and music experts via a stitching-of-experts approach to jointly generate synchronized audio and video \cite{wang2025universe1}.  
\textbf{ThinkSound} frames audio generation as a reasoning process via chain-of-thought, decomposing generation into stages of Foley synthesis, object-centric refinement, and editing, guided by a multimodal LLM \cite{liu2025thinksound}.  
More recently, \textbf{DeepSound-V1} also introduces stepwise CoT reasoning in video→audio synthesis \cite{liang2025deepsoundv1}, and \textbf{YingSound} uses a multimodal CoT controller plus conditional flow matching for sound effect generation in few-shot settings \cite{chen2024yingsound}.  
These works complement ours: while they may retrain large joint models or adopt reasoning-based pipelines, our approach uniquely freezes a strong text–audio backbone and learns only a light cross-modal bridge for alignment.

%\paragraph{Positioning of our approach.}
%Rather than bolting controllers onto a T2A backbone~\cite{zhang2024foleycrafter} or retraining a fully multimodal stack end-to-end~\cite{cheng2024mmaudio,shan2025hunyuanfoley}, we take a middle path: keep the audio prior intact and learn only a thin interface that lets video nudge it. Concretely, the VAE, DiT, and text pathway remain frozen; we tune only lightweight \emph{collaboration layers} and \emph{alignment heads} that read V-JEPA2 tokens. Token-drop and a cached video state bound compute, while RoPE supplies ordering cues without touching backbone weights. This division of labor yields competitive synchrony with far fewer trainable parameters and memory, and—crucially—preserves the primacy of text prompting: prompts set the scene; video sharpens timing and local dynamics.

\subsection{Architectural Positioning of our approach}

Our approach differs from prior adapter-based frameworks for adding multi-modality to frozen single modality models such as FoleyCrafter~\cite{zhang2024foleycrafter} or Stylecodes\cite{rowles2024stylecodes}, which attach specialized controllers to a U-Net backbone for alignment. 
While such modular control can improve synchronization under limited data, it partitions the conditioning pathways -- forcing each module to learn its own alignment rather than leveraging the pretrained model’s holistic structure.

In contrast, \textbf{Foley Control} adopts a more unified transformer-based design that integrates video conditioning directly within the frozen diffusion transformer’s existing attention layers.  
This avoids separate control heads and allows cross-modal signals to propagate through the same representational channels as text, better aligning with modern large-scale pretrained architectures such as Stable Audio Open~\cite{evans2024stableaudioopen}.  
Results from large-scale modeling~\cite{kaplan2020scalinglawsneurallanguage} indicate that architectures which enable pretrained components to co-adapt through shared attention tend to harness scale and generalize more effectively than systems with manually partitioned control modules.  

At the other end of the spectrum, fully multi-modal diffusion transformers such as MMAudio~\cite{cheng2024mmaudio} and HunyuanVideo-Foley~\cite{shan2025hunyuanfoley} extend this idea further by training end-to-end across text, video, and audio streams.  
These models demonstrate even higher efficiency and expressivity when massive, curated datasets are available, but they require orders of magnitude more paired data and compute to converge.  
\textbf{Foley Control} therefore strikes a middle ground: it retains the scalability and representational advantages of transformer conditioning while remaining data-efficient by freezing the text–audio prior and learning only lightweight video–audio bridges.

\section{Method}
\label{sec:method}

\subsection{Preliminaries}

Our approach builds upon two key components: a frozen audio generation backbone and pretrained video encoders for semantic grounding.  
We briefly review the necessary background.

\paragraph{Audio Latent Diffusion.}  
We adopt the \textbf{Stable Audio DiT}~\cite{evans2024stableaudioopen} as the generative backbone.  
Stable Audio is a diffusion-based model operating in the latent space of an audio autoencoder, enabling high-fidelity waveform synthesis at a sampling rate of 44.1\,kHz.  
Given conditioning embeddings (e.g., text or duration), the model learns to denoise latent audio representations over a fixed number of timesteps.  
In our framework, the backbone remains \emph{fully frozen}, ensuring training efficiency and stability.

\paragraph{Video Representation Learning.}  
For visual grounding, we leverage \textbf{V-JEPA2}~\cite{assran2025vjepa2}, a transformer-based video encoder pretrained with predictive objectives.  
Given a sequence of frames, V-JEPA2 produces spatiotemporal patch-level embeddings, which can be pooled into \emph{tubelets} or spatial grids (e.g., $4{\times}4$, $8{\times}8$) to capture both global dynamics and localized motion cues.  
These embeddings serve as \emph{key} tokens for cross-modal alignment with the audio latent sequence.

\paragraph{Problem Setup.}  
Formally, given a video segment $\mathcal{V} = \{f_1, \ldots, f_T\}$, our goal is to synthesize an aligned audio waveform $x \in \mathbb{R}^{L}$, where $L$ corresponds to the clip duration at 44.1\,kHz.
The video encoder maps $\mathcal{V}$ to a sequence of tokens $\mathbf{v} \in \mathbb{R}^{S_v \times D_v}$, while the audio diffusion model operates on latent sequences $\mathbf{a} \in \mathbb{R}^{S_a \times D_a}$.

\subsection{Dataset Curation}
\label{sec:dataset}

Training high-quality video-to-audio models requires large-scale, temporally aligned multimodal data.  
To this end, we constructed a dataset derived from the \textbf{Kinetics-700} dataset \cite{kinetics700} , which provides a diverse set of human action videos in a wide range of everyday activities.  
Since not all videos contain meaningful or relevant sound events, we applied a data curation pipeline similar to that used by HunyuanVideo-Foley \cite{shan2025hunyuanfoley} . Since the dataset was already partitioned into clips, we first filtered out any silent samples from the dataset, we then used ImageBind~\cite{girdhar2023imagebind} and Meta Audiobox Aesthetics~\cite{tjandra2025meta} scores to filter out both low quality and conceptually distinct samples, ensuring high-fidelity and semantically consistent audio--video pairs similar to the filtering strategy of HunyuanVideo-Foley~\cite{shan2025hunyuanfoley}.

\subsection{Framework Overview}

Our framework is based on \textbf{Stable Audio Open\cite{evans2024stableaudioopen}}, a diffusion transformer (DiT) model for high-fidelity text-to-audio generation.  
Raw waveforms $x \in \mathbb{R}^{L}$ are encoded by an audio VAE into latent sequences $\mathbf{a} \in \mathbb{R}^{S_a \times D_a}$, where $S_a$ is the sequence length and $D_a$ the latent dimensionality.  
The DiT backbone performs \emph{latent diffusion} using the \emph{v-prediction} parameterization: at a random step $t$, we corrupt the clean latent $\mathbf{a}_0$ to $\mathbf{a}_t$ and train the model $v_\theta(\mathbf{a}_t, t, \text{cond})$ to predict the velocity that guides $\mathbf{a}_t$ back toward $\mathbf{a}_0$.  
At inference, we integrate the sampler using the predicted velocities to recover $\mathbf{a}_0$, then decode to waveform via the VAE.

\paragraph{Diffusion Transformer (DiT).}  
At its core, Stable Audio Open employs a stack of transformer blocks designed for sequence modeling in the latent space.  
Each block incorporates multi-head self-attention, feed-forward networks, and cross-attention. 
Text embeddings, obtained from a pretrained T5 encoder~\cite{T5}, are injected through cross-attention, enabling semantic control over the generated audio.  
This design allows fast parallel sampling and supports long-context audio generation at 44.1\,kHz.

\paragraph{Freezing Strategy.}  
In our framework, the entire Stable Audio Open backbone—including the VAE encoder/decoder, DiT blocks, and T5 conditioning layers—remains \emph{frozen}. This design choice ensures stable optimization, reduces computational cost, and preserves the strong generative prior acquired from large-scale audio–text pretraining. 

\begin{figure}
    \centering
    \includegraphics[width=1\linewidth]{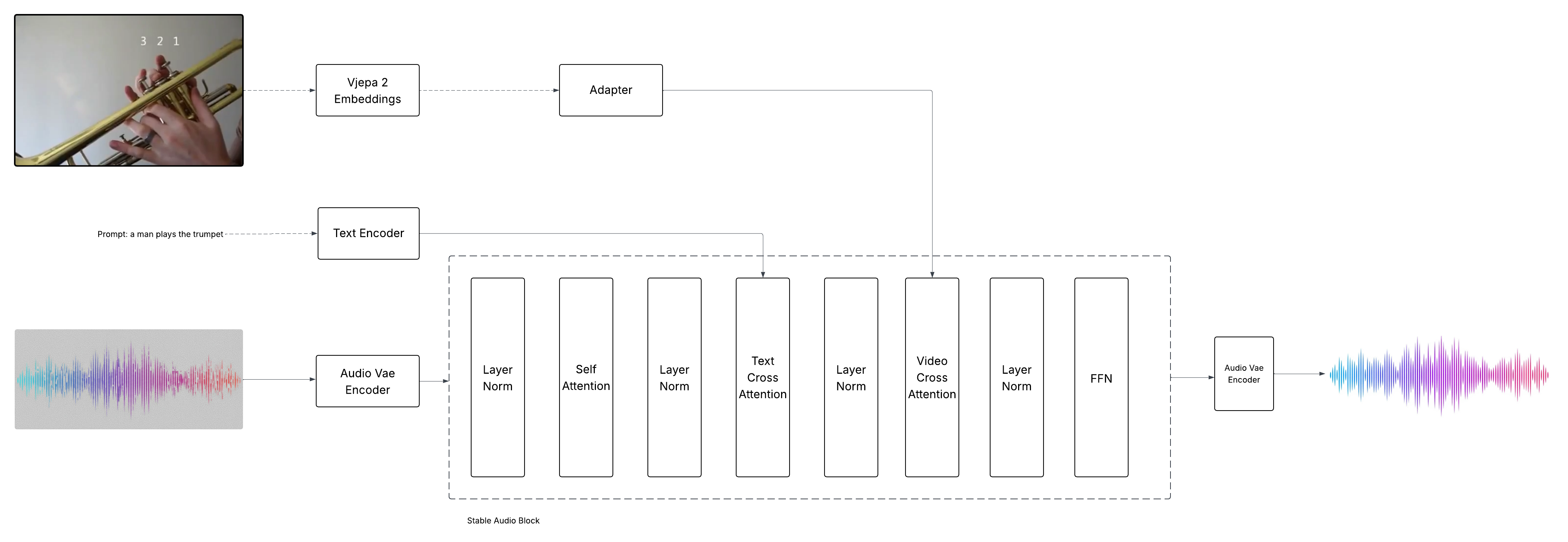}
    \caption{Forward flow: frames $\rightarrow$ V\textendash JEPA2 $\rightarrow$ adapter (video vCA), prompt $\rightarrow$ text encoder (Tx\textendash CA), and noise (latent init) entering the DiT at the same level}
    \label{fig:placeholder}
\end{figure}

\paragraph{Additional Cross-Attention Layers}
To integrate video semantics without disrupting the pretrained frozen stable audio model, we insert video cross-attention \emph{in every DiT block}, immediately after the backbone's text cross-attention and before the feed-forward network (SA $\rightarrow$ Tx-CA $\rightarrow$ \textbf{Vid-CA} $\rightarrow$ FFN).
Audio latents act as queries and video tokens as keys/values; the rest of the block (including the text pathway) remains frozen.

\textbf{Tiny MLP adapter on the video path.}
Before forming $K,V$, we pass the (detached) video features through a lightweight two-layer MLP with GELU and residual addition.

\textbf{RoPE scheme}
We use standard rotary position embeddings (RoPE)~\cite{su2021roformer}.
RoPE is applied independently to audio queries and video keys: each modality computes its own phase from its sequence positions, and the rotations are not shared across modalities.
Concretely, after linear projection we rotate $Q$ and $K$ in-place along their last dimension (per head) prior to attention.
This preserves relative temporal phase information, helping the model align video motion and audio onsets more precisely.
While prior work~\cite{cheng2024mmaudio,synchformer2024iashin} employs specialized synchronization modules (e.g., SyncFormer) for cross-modal alignment, we found RoPE sufficient for stable temporal correspondence without additional alignment networks in our adapter setting.

\textbf{Cross-attention placement.}
Our placement choice is inspired by Kong et al.~\cite{kong2025lettalkaudiodrivenmultiperson}, which augments each DiT block with an audio cross-attention layer inserted after the text cross-attention. 
Following this design, we adopt the same ordering for our video cross-attention so that prompts establish global semantics before modality-specific timing and dynamics are injected. 
Unlike their setup, which employs label-aware RoPE (L-RoPE) to distinguish multiple audio streams, we found standard RoPE sufficient for stable and precise cross-modal alignment in our single-stream video-to-audio configuration.

\textbf{Scope of training.}
Only the parameters introduced by this sublayer are trainable (video MLP adapter, $W_q$, $W_{kv}$, $W_o$, attention weights, and local norms); all backbone weights, the audio VAE, and the text pathway remain frozen.
Unless otherwise stated, each DiT block has its own (non-shared) set of sublayer parameters.

\paragraph{V\textendash JEPA2 Embedding Pooling.}
We condition collaboration cross-attention on V\textendash JEPA2 tokens derived from $16$\,FPS video streams. For each $4$\,s segment, we sample $64$ frames and encode them with V\textendash JEPA2. To obtain a compact sequence, we \emph{pool each effective frame} into a \emph{single} token (the encoder operates with stride~2, so one effective frame corresponds to two input frames). This results in $32$ effective frames per $4$\,s segment and thus $32$ tokens per segment. To bound computational cost, we restrict inputs to a maximum of $12$\,s and concatenate the segment-level embeddings in temporal order. Originally, we experimented with spatial grids such as $8{\times}8$ ($64$ tokens per frame) and $16{\times}16$ ($256$ tokens per frame), but found that reducing to a single pooled token per frame preserved salient spatial context while substantially improving efficiency and stabilizing optimization.

\section{Experiments}

\subsection{Experimental Setup}
We evaluate our proposed joint audio–video fine-tuning framework on the curated Kinetics-700 dataset (Section~\ref{sec:dataset}), using the large filtered corpus for pretraining and the high-quality SFT subset for supervised alignment.  
We train all the models for the experiment with a batch size of 12, using a frozen \texttt{StableAudioDiT} backbone and V-JEPA2 embeddings; only the collaboration layers are updated. We adopt the original Stable Audio Open velocity-prediction training setup and apply token-drop regularization with 10\% probability. For evaluation, we use the Meta Movie Audio Bench test set dataset.

\subsection{Ablation Studies}
\label{sec:ablation}

We analyze the impact of video embedding granularity on model performance through a series of controlled ablations.

\paragraph{Pooling Strategies.}
We compare two ways of aggregating V-JEPA2 patch tokens into video tokens:
(i) \emph{frame pooling} (1 token per two frames), 
(ii) \emph{grid8} pooling ($8{\times}8$ tokens per frame, 64 per frame).  
Frame pooling offers maximum computational efficiency, while grid-based schemes capture richer spatial and motion cues at higher cost.

We report the KL-PANNs metric computed between generated and ground-truth audio event posteriors on the MovieGenBench test set, without text prompts, to isolate the effect of visual conditioning.

To reduce compute, all ablation runs are trained on a fixed 30\% random subset of our curated Kinetics\textendash700 training split; the same subset is used for both pooling variants, with identical hyperparameters , schedules and seed across conditions.

\begin{table}[h]
\centering
\caption{Ablation study comparing pooling strategies over training steps 
using the KL-PANNs metric (lower is better) on the Kinetics-700 validation 
subset without text guidance.}
\begin{tabular}{lcc}
\hline
\textbf{Training Steps} & \textbf{Grid8} & \textbf{Single Pooled Embedding} \\
\hline
50{,}000  & 3.220921 & 3.222953 \\
100{,}000 & 3.145059 & 3.188678 \\
200{,}000 & 3.153171 & 3.194564 \\
300{,}000 & 3.104110 & 3.130133 \\
400{,}000 & 3.119460 & 3.111351 \\
\hline
\end{tabular}
\label{tab:ablations}
\end{table}

\begin{figure}[htbp]
\centering
\begin{tikzpicture}
\begin{axis}[
    width=0.9\linewidth,
    height=4.5cm,                % ↓ was 6cm, shrunk vertically
    xlabel={Training Steps},
    ylabel={Metric (lower is better)},
    xmin=40000, xmax=420000,     % ↓ start near 0.4e5 instead of 0
    ymin=3.08, ymax=3.25,
    xtick={50000,100000,200000,300000,400000},
    ytick={3.10,3.15,3.20,3.25},
    legend style={at={(0.5,1.02)},anchor=south,legend columns=-1},
    legend pos=north east,
    grid=major,
    grid style={dashed,gray!20},
    thick
]

\addplot[color=orange, very thick, mark=square*] coordinates {
    (50000,3.220921)
    (100000,3.145059)
    (200000,3.153171)
    (300000,3.104110)
    (400000,3.119460)
};
\addlegendentry{grid8}

\addplot[color=gray, dashed, mark=triangle*] coordinates {
    (50000,3.222953)
    (100000,3.188678)
    (200000,3.194564)
    (300000,3.130133)
    (400000,3.111351)
};
\addlegendentry{Single Pooled}

\end{axis}
\end{tikzpicture}
\caption{Ablation trends across training steps for different pooling strategies, 
evaluated with the KL-PANNs metric (lower is better) on the Kinetics-700 validation 
subset without text guidance. All runs were trained on a fixed 30\% subset of the 
curated training data.}
\label{fig:abl_plot}
\end{figure}
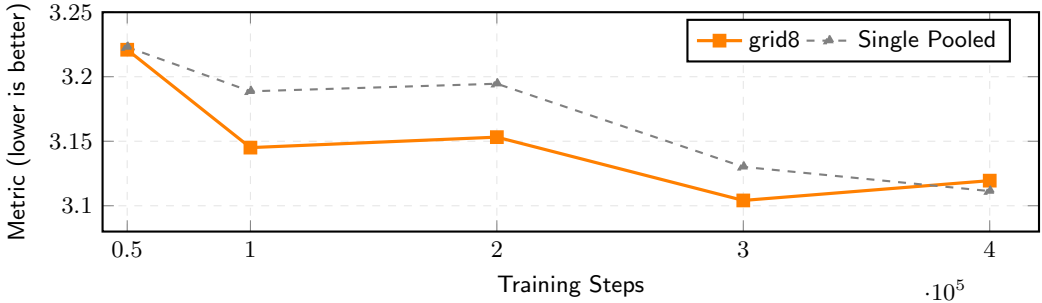

\paragraph{Results and Discussion.}
As shown in Table~\ref{tab:ablations} and Figure~\ref{fig:abl_plot}, the lower-resolution \textbf{Single pooled embedding} configuration achieves performance on par with, or slightly better than, the grid8 embedding variant.  
Despite a substantial reduction in visual token count---and therefore compute and memory use---no meaningful loss in temporal alignment or perceptual fidelity was observed.  
Across 400k training steps, the metrics differ by less than $0.03$, indicating that the mid-level spatial resolution of the single-pooled embeddings captures sufficient motion and context cues for Foley synchronization.

\subsection{Comparison to Existing Work}
\label{sec:sota}
We compare our framework against recent state-of-the-art video-to-audio generation systems, including MMAudio~\cite{cheng2024mmaudio}, HunyuanVideo-Foley~\cite{shan2025hunyuanfoley}, ThinkSound, and FRIEREN.  
All baselines represent distinct strategies for bridging video and audio modalities, ranging from fully joint multimodal diffusion training to modular adapter-based control.  
To ensure comparability, we evaluated all models under a consistent protocol using the \textbf{MovieGenBench}\cite{polyak2024movie} benchmark, which emphasizes long-form cinematic scenes with diverse dynamics and complex soundscapes.

\paragraph{Evaluation Protocol.}
Each method generates audio at 44.1\,kHz, conditioned on video frames and corresponding text prompts when supported.  
For our model, the Stable Audio DiT backbone, VAE, and CLAP text encoder remain entirely \emph{frozen}; only the lightweight collaboration layers and alignment heads introduced in Section~\ref{sec:method} are trained.  
This setup isolates the effect of our proposed video bridge while maintaining a fixed generative prior across all experiments.  
All systems are evaluated on the same MovieGenBench test split, using synchronized video clips with corresponding ground-truth soundtracks. 
\textbf{For metric computation, preprocessing, and dataset loaders, we used the MMAudio evaluation/testing repository}\footnote{\url{https://github.com/facebookresearch/mmaudio/tree/main/eval}} \textbf{to ensure consistent scoring across methods}.

\paragraph{Metrics.}
Following the official benchmarking suite~\cite{shan2025hunyuanfoley}, we report a comprehensive set of perceptual and statistical metrics capturing complementary aspects of generation quality, including Fréchet Distance and KL divergence using PANNs~\cite{kong2020panns} and PaSST~\cite{koutini22passt}, cross-modal consistency via ImageBind~\cite{girdhar2023imagebind} and synchronization via Synchformer~\cite{synchformer2024iashin}

\begin{itemize}
\item \textbf{ImageBind Score (IB)} quantifying audio–visual semantic consistency via cosine similarity between audio and image frame embeddings extracted by \textbf{ImageBind}\cite{girdhar2023imagebind}.
\item \textbf{Mean KL Divergence (KL)} between classifier-based audio event posteriors, using PaSST (\textbf{KL-PaSST}\cite{koutini22passt}) and PANNs (\textbf{KL-PANNs}). Lower values denote better distributional consistency.
\item \textbf{Fréchet Distance (FD)} between generated and real audio embeddings, computed with three pretrained encoders: PaSST (\textbf{FD-PaSST}), PANNs (\textbf{FD-PANNs}), and VGGish (\textbf{FD-VGG}). Lower values indicate closer alignment between the generated and reference distributions.
\item \textbf{DeSync Score} evaluating temporal misalignment (in seconds) predicted by \textbf{Synchformer\cite{synchformer2024iashin}}; lower values indicate better synchronization.
\end{itemize}

\paragraph{Baselines.}
\begin{itemize}
    \item \textbf{MMAudio}~\cite{cheng2024mmaudio}: a fully multimodal diffusion transformer jointly trained on text, audio, and video under a conditional flow-matching objective.
    \item \textbf{HunyuanVideo-Foley}~\cite{shan2025hunyuanfoley}: a large-scale multimodal DiT trained end-to-end on curated text--video--audio data, incorporating representation-alignment losses for temporal precision.
    \item \textbf{ThinkSound}~\cite{liu2025thinksound}: a modular system combining CoT reasoning with pretrained encoders and a controllable diffusion backbone, designed for robustness to domain variation.
    \item \textbf{FRIEREN}~\cite{wang2024frieren}: an autoregressive video-to-sound system emphasizing temporal causality and synchronization through hierarchical attention mechanisms.
    \item \textbf{FoleyCrafter}~\cite{zhang2024foleycrafter}: an adapter-based approach that enhances synchronization by injecting semantic and temporal controllers into a pretrained text-to-audio backbone, improving alignment without retraining large generative models.
\end{itemize}

\paragraph{Fairness and Implementation Details.}
All comparisons use identical input video frame rates and duration limits.  
During evaluation, each model produces a single audio sample per clip without post-processing, ensuring consistency across systems and avoiding any external enhancement or mixing effects.  
We do not include comparisons against \textbf{V-AURA}~\cite{viertola2025temporally}, as the method is constrained to clips of 2.5\,seconds in duration, which makes it unsuitable for evaluation on longer-form datasets such as MovieGenBench that emphasize multi-second temporal dependencies and ambient context.

\paragraph{MovieGenBench.}
We test on the \textbf{MovieGenBench} dataset~\cite{polyak2024movie}, which emphasizes cinematic sound design and long-range temporal dependencies.  
This benchmark evaluates generated audio against a strong text-to-video-with-audio (T2VA) reference model, providing a measure better aligned with large-scale multimodal systems such as MMAudio~\cite{cheng2024mmaudio} and HunyuanVideo-Foley~\cite{shan2025hunyuanfoley}, which excel at generating ambient, scene-level audio.  
Because MovieGenBench includes extensive background and environmental textures, it favors models that maintain coherent ambiance and long-horizon consistency rather than isolated transients.  
In contrast, \textbf{VGGSound}~\cite{Chen20} consists primarily of short, event-driven Foley-style clips that emphasize localized synchronization and sound event accuracy.  
We also omit comparisons with \textbf{V-AURA}~\cite{viertola2025temporally}, a video-to-audio model limited to generating 2.5-second clips, which makes it unsuitable for long-form benchmarks like MovieGenBench.  
As shown in Table~\ref{tab:comparison-movie}, Foley Control performs competitively under these more demanding, ambient conditions.

\paragraph{Kling-Foley AudioEval.}
To avoid train–evaluation contamination, we do not report results on the Kling-Foley AudioEval benchmark introduced by Wang et~al.~\cite{wang2025klingfoley}.  
Our training corpus included material overlapping or closely related to that evaluation split, which could yield inflated or non-comparable scores.

\begin{table}[h]
\centering
\footnotesize
\caption{Comparison on the MovieGenBench dataset.}
\setlength{\tabcolsep}{4pt}
\begin{tabular}{lcccccccc}
\hline
\textbf{System} &
\textbf{KL-\scalebox{0.7}{PANNs} ↓} &
\textbf{KL-\scalebox{0.7}{PaSST} ↓} &
\textbf{IB ↑} &
\textbf{FD-\scalebox{0.7}{VGG} ↓} &
\textbf{FD-\scalebox{0.7}{PANNs} ↓} &
\textbf{FD-\scalebox{0.7}{PaSST} ↓} &
\textbf{DeSync ↓} \\
\hline
\textbf{FRIEREN}       & 3.58 & 3.89 & 0.14 & 5.65 & 59.04 & 560.91 & 0.30 \\
\textbf{MMaudio}       & 2.52 & 2.35 & 0.25 & \textbf{4.14} & 37.60 & \textbf{343.24} & \textbf{0.29} \\
\textbf{HunyuanVideo-Foley} & 2.58 & 2.11 & \textbf{0.30} & 7.00 & 31.28 & 373.62 & 0.31 \\
\textbf{Foley Control (ours)} & 2.93 & 2.59 & 0.20 & 5.89 & \textbf{31.10} & 383.99 & 0.32 \\
\textbf{ThinkSound}    & 3.16 & 2.90 & 0.18 & 6.62 & 33.62 & 468.25 & 0.30 \\
\textbf{FoleyCrafter}  & \textbf{1.11} & \textbf{1.29} & 0.26 & 6.94 & 40.70 & 493.08 & 0.33 \\
\hline
\end{tabular}
\label{tab:comparison-movie}
\end{table}

\paragraph{Training Efficiency.}
While large multimodal diffusion systems such as HunyuanVideo-Foley train end-to-end for \textbf{200k--700k steps} on roughly \textbf{100k hours} of curated text--video--audio data using \textbf{128$\times$H20 GPUs} and an \textbf{effective batch size of 2048}, our Foley Control bridge trains for only \textbf{400k steps} with an \textbf{effective batch size of 384}. 
In contrast to MMAudio and ThinkSound, which use a comparable amount of paired audio--video data but additionally rely on extensive \emph{audio-only} pretraining to learn their generative priors, Foley Control requires \emph{no} such auxiliary corpus—leveraging instead a frozen Stable Audio backbone trained independently on text--audio data. 
Compared to HunyuanVideo-Foley, Foley Control operates with nearly \textbf{two orders of magnitude less paired data and compute}, yet achieves competitive synchronization and semantic alignment, underscoring the efficiency of the lightweight cross-modal adapter strategy.
\emph{Similarly, FoleyCrafter}~\cite{zhang2024foleycrafter} demonstrates that adapter-based designs can deliver strong alignment and controllability by keeping a pretrained T2A backbone frozen and learning only compact temporal and semantic controllers, employing a more elaborate adapter architecture built atop a U-Net–based generative model.

\section{Conclusion}

We introduced \textbf{Foley Control}, a lightweight bridge that brings video guidance to a frozen text-to-audio generator by inserting compact, trainable collaboration layers after the model’s existing text cross-attention. With V-JEPA2 embeddings, token pooling, and RoPE-based ordering cues, our design preserves the strengths of the audio prior and prompt controllability while adding the temporal control needed for Foley.

Across a curated data corpus and evaluation on MovieGenBench, the approach delivers competitive semantic and temporal alignment while training only a small fraction of parameters compared to fully multimodal systems. Ablations show that aggressively pooled video tokens match the performance of denser grid features, substantially reducing compute and memory without degrading synchronization.

Practically, the framework remains modular: encoders or the T2A backbone can be swapped or upgraded without end-to-end retraining, which is attractive for production settings where models evolve, additional control modules are added and latency/VRAM budgets matter.

\paragraph{Limitations and Future Work.}
Our current setup caps video duration and conditions on pooled tokens, which may miss rare fine-grained spatial cues. The method also assumes clean inputs and does not explicitly model spatial (binaural/ambisonic) acoustics or streaming/online alignment. Future work may include adaptive tokenization (learned pooling or budget-aware routing), longer-context conditioning, more varied data, spatial audio generation, robustness to in-the-wild edits and background music, and extending the bridge to other audio modalities such as speech and dialogue.

\bibliographystyle{plain}
\bibliography{sample}

\end{document}